\documentclass[runningheads]{llncs}
\usepackage{graphicx}
\usepackage{todonotes}

\newcommand\green[2]{\textcolor{green}{#2}}
\usepackage{tabularx}
\usepackage{pifont}
\newcommand{\cmark}{\ding{51}}%
\usepackage{amsmath}
\usepackage{amssymb}
\usepackage{booktabs}

\usepackage[acronym]{glossaries}
\newacronym{ALS}{ALS}{airborne laser scanning}
\newacronym{MLS}{MLS}{mobile laser scanning}
\newacronym{LoD}{LoD}{level of detail}
\newacronym{OGC}{OGC}{Open Geospatial Consortium}
\newacronym{GML}{GML}{Geography Markup Language}
\newacronym{ASAM}{ASAM}{Association for Standardization of Automation and Measuring Systems}
\newacronym{TLS}{TLS}{terrestrial laser scanning}
\newacronym{UAV}{UAV}{unmanned aerial vehicle}
\newacronym{HD}{HD}{high definition}
\newacronym{RANSAC}{RANSAC}{RANdom SAmple Consensus}
\newacronym{ROI}{ROI}{region of interest}
\newacronym{DEM}{DEM}{digital elevation model}
\newacronym{ICP}{ICP}{iterative closest point}
\newacronym{NLOS}{NLOS}{non-line-of-sight}
\newacronym{SfM}{SfM}{structure from motion}
\newacronym{FME}{FME}{Feature Manipulation Engine}
\newacronym{OSM}{OSM}{OpenStreetMap} 
\newacronym{RMSE}{RMSE}{root mean square error}
\newacronym{CPT}{CPT}{conditional probability table}
\newacronym{DST}{DST}{Dempster–Shafer theory}
\newacronym{BN}{BayNet}{Bayesian network}
\newacronym{GIS}{GIS}{Geographic Information System}
\newacronym{PPD}{PPD}{posterior probability distribution}
\newacronym{CI}{CI}{confidence interval}

\begin{document}
%
%
\title{MLS2LoD3: Refining low LoDs building models with MLS point clouds to reconstruct semantic LoD3 building models}
\titlerunning{MLS2LoD3}

\author{Olaf Wysocki\textsuperscript{1 }, Ludwig Hoegner\textsuperscript{1,2 }, Uwe Stilla\textsuperscript{1 }}
\authorrunning{Wysocki O., et al.,}
%
 \institute{Photogrammetry and Remote Sensing, TUM School of Engineering and Design, Technical University of Munich (TUM),\\ Munich, Germany - (olaf.wysocki, ludwig.hoegner, stilla)@tum.de \and Department of Geoinformatics, University of Applied Science (HM), Munich, Germany - ludwig.hoegner@hm.edu\\}

\maketitle              
\begin{abstract}
Although highly-detailed LoD3 building models reveal great potential in various applications, they have yet to be available.
The primary challenges in creating such models concern not only automatic detection and reconstruction but also standard-consistent modeling. 
In this paper, we introduce a novel refinement strategy enabling LoD3 reconstruction by leveraging the ubiquity of lower LoD building models and the accuracy of MLS point clouds. 
Such a strategy promises at-scale LoD3 reconstruction and unlocks LoD3 applications, which we also describe and illustrate in this paper.
Additionally, we present guidelines for reconstructing LoD3 facade elements and their embedding into the CityGML standard model, disseminating gained knowledge to academics and professionals.
We believe that our method can foster development of LoD3 reconstruction algorithms and subsequently enable their wider adoption.

\end{abstract}
\section{Introduction}
Reconstructing semantic 3D building models is a long-standing challenge in geoinformatics and photogrammetry.
Recent open source and proprietary solutions have proved effective for country-scale~\gls{LoD} 1 and 2 reconstructions~\footnote{https://github.com/OloOcki/awesome-citygml}, given that aerial observations and footprints are provided~\cite{RoschlaubBatscheider}.
Such 3D models find their applications in various fields ranging from analysing roof solar potential to estimating building-energy demand~\cite{biljeckiApplications3DCity2015}.

However, at-scale reconstruction of semantic 3D building models at~\gls{LoD}3 remains in its infancy~\cite{biljecki2014formalisation}. 
The primary obstacle has been the lack of highly-detailed, street-level data capturing the distinctive feature of the~\gls{LoD}3: facade elements. 
Numerous initiatives stemming from academia and industry promise to close this data-gap by conducting large-scale \gls{MLS} campaigns acquiring high-density, street-level point clouds and images~\cite{tumfacadePaper}.

Much research has been devoted to reconstructing facade elements~\cite{szeliski2010computer}, yet methods embedding these elements directly into the semantic 3D building models standards' structure are scarce.
The principal challenges lie in the incompleteness of acquired \gls{MLS} data, where often only the frontal facade is measured.
This feature hinders from-scratch reconstruction, which assumes very high point cloud coverage~\cite{pantoja2022generating,helmutMayerLoD3,nan2017polyfit}.
Another challenge concerns single-object semantics, typically derived in the semantic segmentation process. 
Such an approach implicitly discards the hierarchical complexity of the semantic data model, rendering it insufficient for directly creating semantic 3D data models, for instance, embedding 3D window objects into a wall surface belonging to a building entity within a city model~\cite{beil2021integration}.

In this paper, as opposed to traditional from-scratch reconstruction, we propose the refinement strategy to create~\gls{LoD}3 building models.
The strategy minimizes the influence of data incompleteness by
using low~\gls{LoD} solid geometry as prior and adding new geometry only where required.
The method also reduces the complexity of maintaining hierarchical reconstruction by embedding new objects into existing model structures;
We list our contributions as follows:
\begin{itemize}
    \item Conflict-driven refinement strategy of \gls{LoD} 1 and 2 to \gls{LoD}3 building models while considering uncertainty
    \item Introducing guidelines of semantic~\gls{LoD}3-level facade elements reconstruction
    \item Presenting applications of~\gls{LoD}3 building models
\end{itemize}

\section{LoD3 applications}
\begin{figure}
\includegraphics[width=\textwidth]{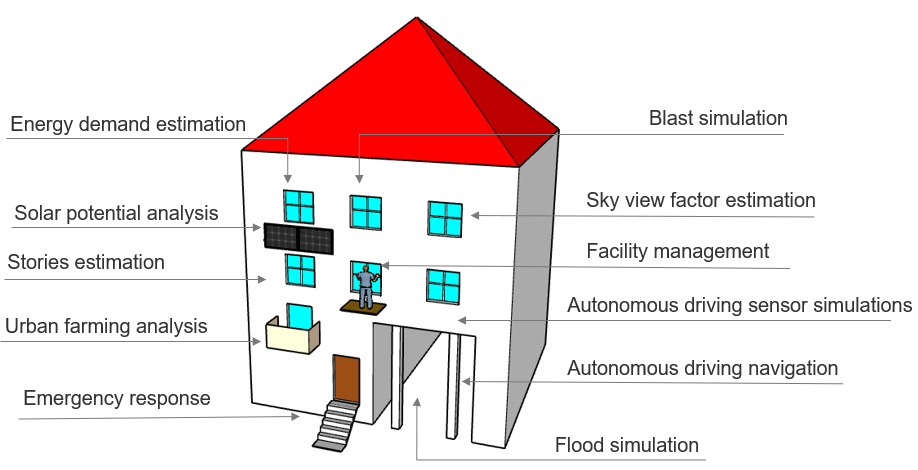}
\caption{Applications of semantic 3D building models at~\gls{LoD}3} 
\label{applications}
\end{figure}
There are multiple studies elaborating on applications of semantic 3D building models, which, however, concentrate on~\gls{LoD} 1 and 2 models \cite{biljeckiApplications3DCity2015,willenborg2018applications}.
This phenomenon primarily owes to the ubiquity of \gls{LoD} 1 and 2 models; for example, there are approximately 140 million freely available semantic 3D building models in the US, Switzerland, and Poland~\footnote{https://github.com/OloOcki/awesome-citygml}.

However, many professionals and academics have already expressed the need for~\gls{LoD}3 building models in various applications.
As illustrated in Figure~\ref{applications}, the applications concern simulating flood damage~\cite{amirebrahimi2016bim}, estimating heating demand~\cite{nouvel2013citygml}, calculating façade solar potential~\cite{willenborgIntegration2018}, testing automated driving functions~\cite{schwabRequirementAnalysis3d2019}, supporting autonomous driving navigation~\cite{wysockiUnderpasses}, managing building facilities~\cite{moshrefzadeh2015citygml}, measuring sky-view factor~\cite{lee2013modelling}, analysing building potential for vertical farming~\cite{palliwal20213d}, investigating blasts impact~\cite{willenborg2016semantic}, navigating for emergency units \cite{kolbe2008citygml}, and inferring number of stories \cite{biljecki2021street}.
\section{Refining LoD 1 and 2 to obtain LoD3 building models}
\begin{figure}
\includegraphics[width=\textwidth]{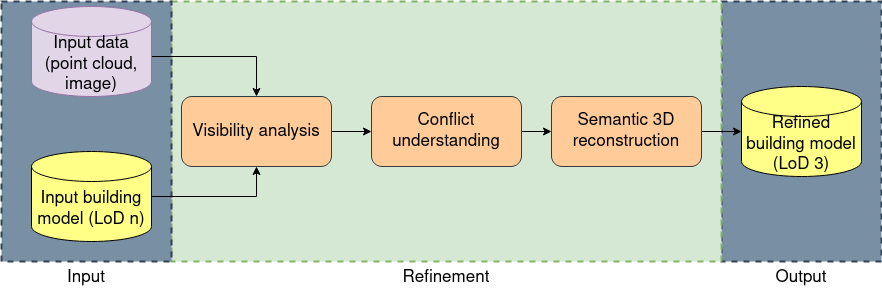}
\caption{Refinement strategy workflow} 
\label{overview}
\end{figure}
\begin{figure}
\includegraphics[width=\textwidth]{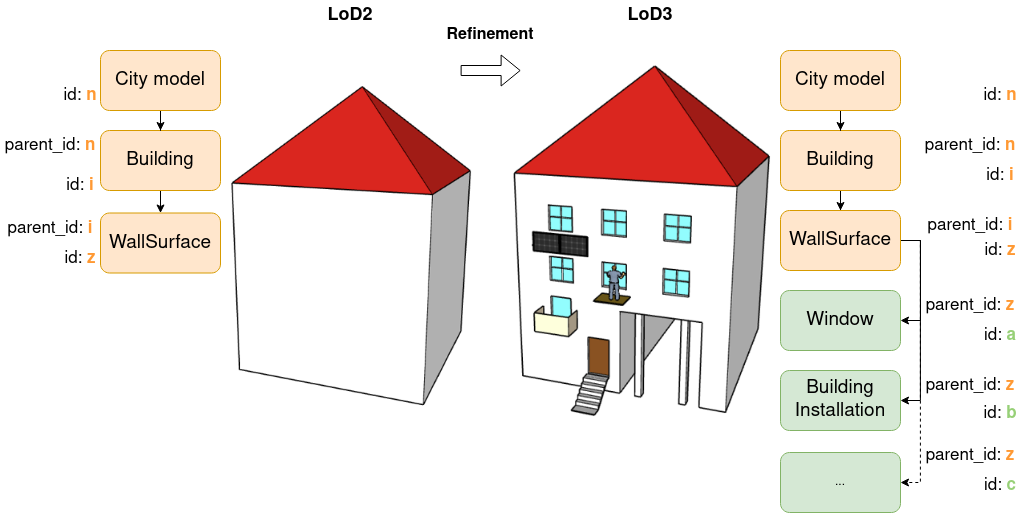}
\caption{Concept of the refinement: Wall Surface serves as a projection plane for facade elements and as a link to building entity and subsequently city model.} 
\label{refinementEmbed}
\end{figure}
%
As shown in the strategy overview in Figure~\ref{overview}, the measurements and 3D building models are analyzed in the visibility analysis part, which estimates absent objects in~\gls{LoD} 1 or 2 (conflicts).
The conflict semantic is derived in the following step. 
The semantic 3D reconstruction part not only reconstructs geometries but also assigns them semantics and embeds into the standardized, CityGML 2.0 building model representation~\cite{grogerOGCCityGeography2012}; yielding refined, \gls{LoD}3 building model.
Since the visibility analysis and conflict understanding parts are at length presented in publications such as~\cite{wysockiUnderpasses,wysockiVisibility,wysocki2023scan2lod3,tuttas_reconstruction_2013,tuttas2015validation}.
We present merely an excerpt about them in Section~\ref{visibilityanalysis}, and elaborate on embedding reconstructed objects into the widely-adopted CityGML standard of semantic 3D building models in Section~\ref{embeding}.
The concept of the refinement exploits planar surfaces present in the semantic 3D city models both for detection and reconstruction as well as for the semantic embedding into the model, as illustrated in Figure~\ref{refinementEmbed}.
\subsection{Visibility analysis \& Conflict understanding}
\label{visibilityanalysis}
\begin{figure}
\includegraphics[width=\textwidth]{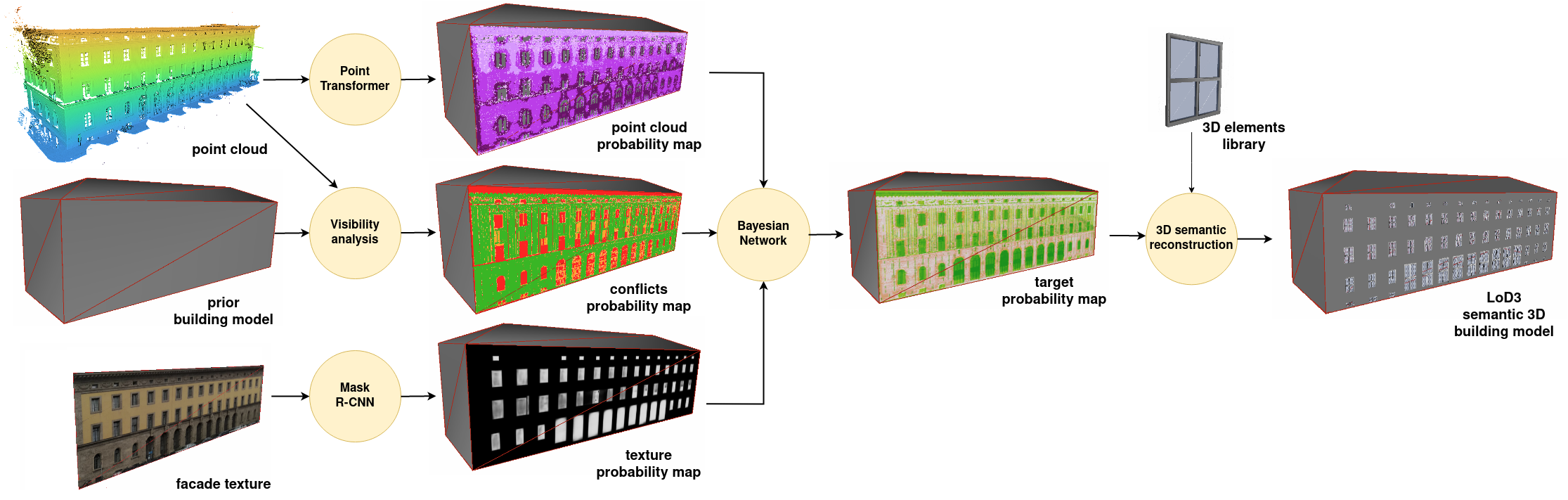}
\caption{Conflicts understanding performed using combination of visibility analysis, semantically segmented point clouds and images. Adapted from~\cite{wysocki2023scan2lod3}.} 
\label{conflictUnderstanding}
\end{figure}
\begin{figure}
\includegraphics[width=\textwidth]{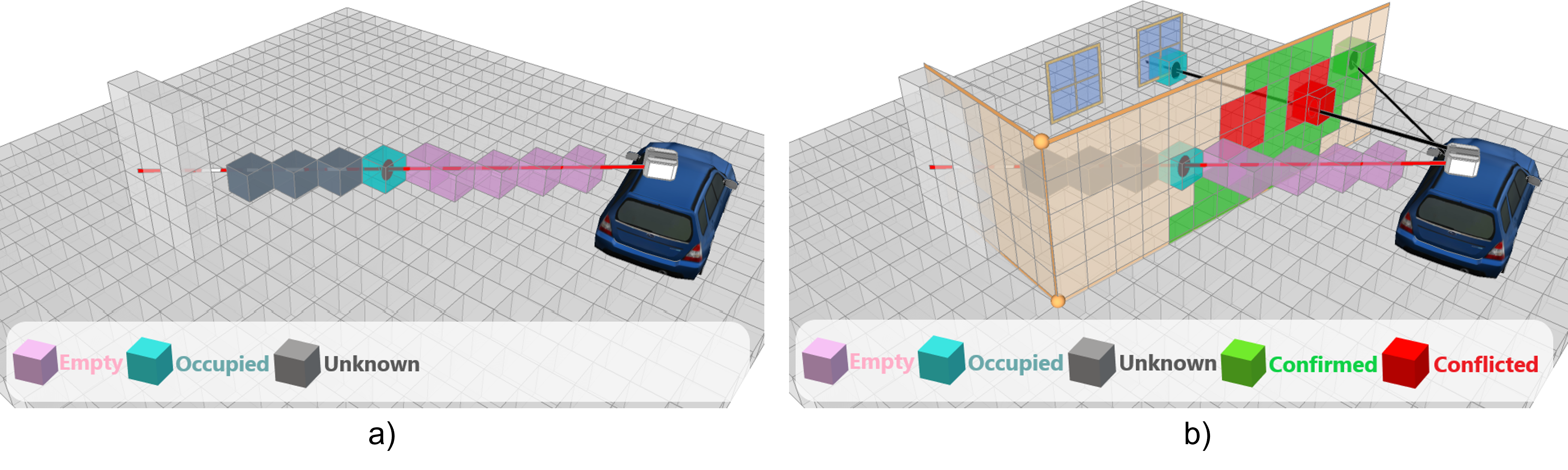}
\caption{Visibility analysis employed on a voxel grid to identify~\gls{LoD}3 objects, absent in~\gls{LoD} 1 and 2 building models (\textit{conflicts}). a) Ray casting from sensor origin provides voxel state \textit{empty} if the observation ray traverses it; \textit{occupied} when it contains hit point; \textit{unknown} if unmeasured; b) Joint analysis of rays and vector models provides another set of states: \textit{confirmed} when \textit{occupied} voxel intersects with vector plane; and \textit{conflicted} when the plane intersects with an \textit{empty} voxel. Adapted from~\cite{wysocki2023scan2lod3}.} 
\label{raycasting}
\end{figure}
The method comprises three main input blocks, as shown in the overview in Figure~\ref{conflictUnderstanding}.
The backbone of our method is the visibility analysis, where conflicts between laser rays and vectors are analyzed and projected as~\textit{conflict probability map} to the wall, yielding geometric cues about missing~\gls{LoD}3 facade elements (Figure~\ref{raycasting}).
Both modalities have different uncertainties of global positioning accuracy of building model surfaces and of point clouds along the ray.
We assume the probability distribution of global positioning accuracy of a building surface $P(A)$ is described by the Gaussian distribution \begin{math} {\mathcal{N}} (\mu_{1}, \sigma_{1}) \end{math}, where $\mu_{1}$ and $\sigma_{1}$ are the
mean and standard deviation of the Gaussian distribution.
Analogically, we describe the probability distribution of global positioning accuracy of a single point in point cloud $P(B)$ by the Gaussian distribution ${\mathcal{N}} (\mu_{2}, \sigma_{2})$.
We use the joint probability distribution of two independent events $P(A)$ and $P(B)$ to obtain final probability scores for the $P_{confirmed}$ and conflicted $P_{conflicted}$ states of the voxel $V_{n}$~\cite{wysocki2023scan2lod3}:
\begin{equation}
V_{n} = 
 \begin{Bmatrix}
  P_{confirmed}(A,B) = P(A) * P(B)\\
  P_{conflicted}(A,B) = 1 - P_{confirmed}(A,B)\\
 \end{Bmatrix} 
\end{equation}

We derive the semantic information about conflicts from two data sources: 3D point clouds and images projected to the wall.
The 3D point clouds are semantically segmented using the enhanced Point Transformer network~\cite{wysocki2023scan2lod3}. 
Their final class scores are projected onto the wall surface, forming a~\textit{point cloud probability map}.
Analogically, \textit{texture probability map} is created from the inferred classes of the Mask-RCNN network~\cite{wysocki2023scan2lod3} applied to the building wall texture. 
The probability maps are combined using the Bayesian network, which estimates geometry and semantics for the facade elements reconstruction.
We automatically reconstruct facade elements by fitting a pre-defined library of objects into their identified respective instances.
The fitting comprises a selection of a corresponding 3D model matched with the detected class and rigid scaling to the identified extent. 

\subsection{Embedding 3D reconstructed objects into the CityGML standard}
\label{embeding}
\begin{figure}
\includegraphics[width=\textwidth]{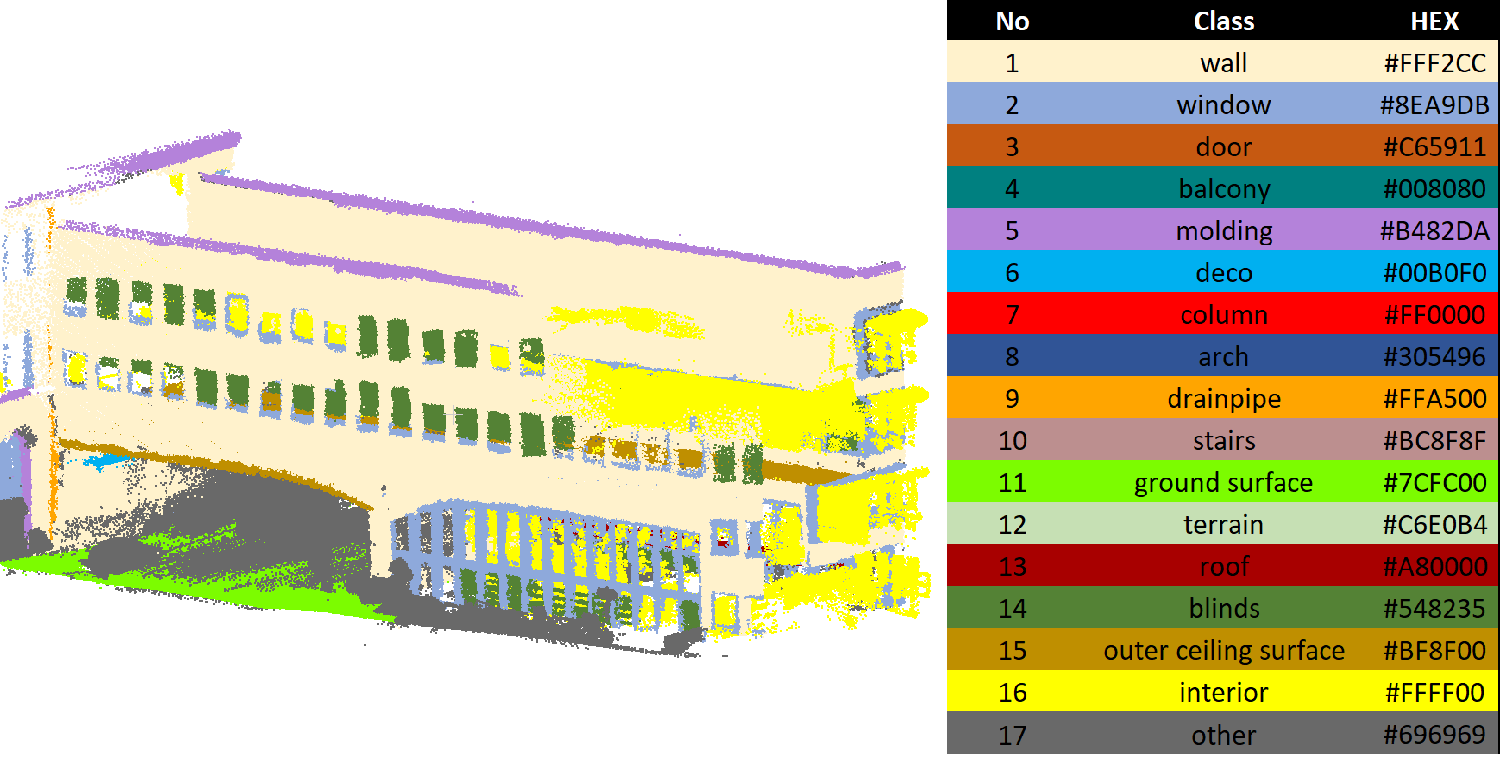}
\caption{Facade-level classes available in point cloud datasets on an example of the TUM-FAÇADE dataset~\cite{tumfacadePaper}.} 
\label{tumfacadeclasses}
\end{figure}

For semantic segmentation of facade-level~\gls{MLS} point clouds, up to 17 classes can be recognized (Figure~\ref{tumfacadeclasses}). 
As we show in Table~\ref{codes}, these are pertinent to the reconstruction adhering to the CityGML 2.0 standard~\cite{grogerOGCCityGeography2012,special_interest_group_3d_modeling_2020,biljecki2014formalisation}. 
However, the refinement necessitates addressing application-specific requirements, prior model's \gls{LoD}, and characteristics of street-level measurements (Table~\ref{codes}).

Ground Surface, Roof Surface, and Wall Surface positions shall not be altered, as it will corrupt the overall geometry of the prior model since the street-level measurement typically covers only one or two sides of the building.
Such change shall be only undertaken if the point cloud significantly covers the building's outer shell (e.g., see~\cite{nan2017polyfit}) and is of higher global accuracy than usually cadastre-derived Wall Surface~\cite{RoschlaubBatscheider}. 
Yet, even when the geometry is altered, the entity's identifiers shall not be changed, as there exist external and internal links associated with city model objects. 
For example, an identifier of a wall can be linked to a report about its solar potential; in case the identifier is changed, this information will be lost. 

\begin{figure}
\includegraphics[width=\textwidth]{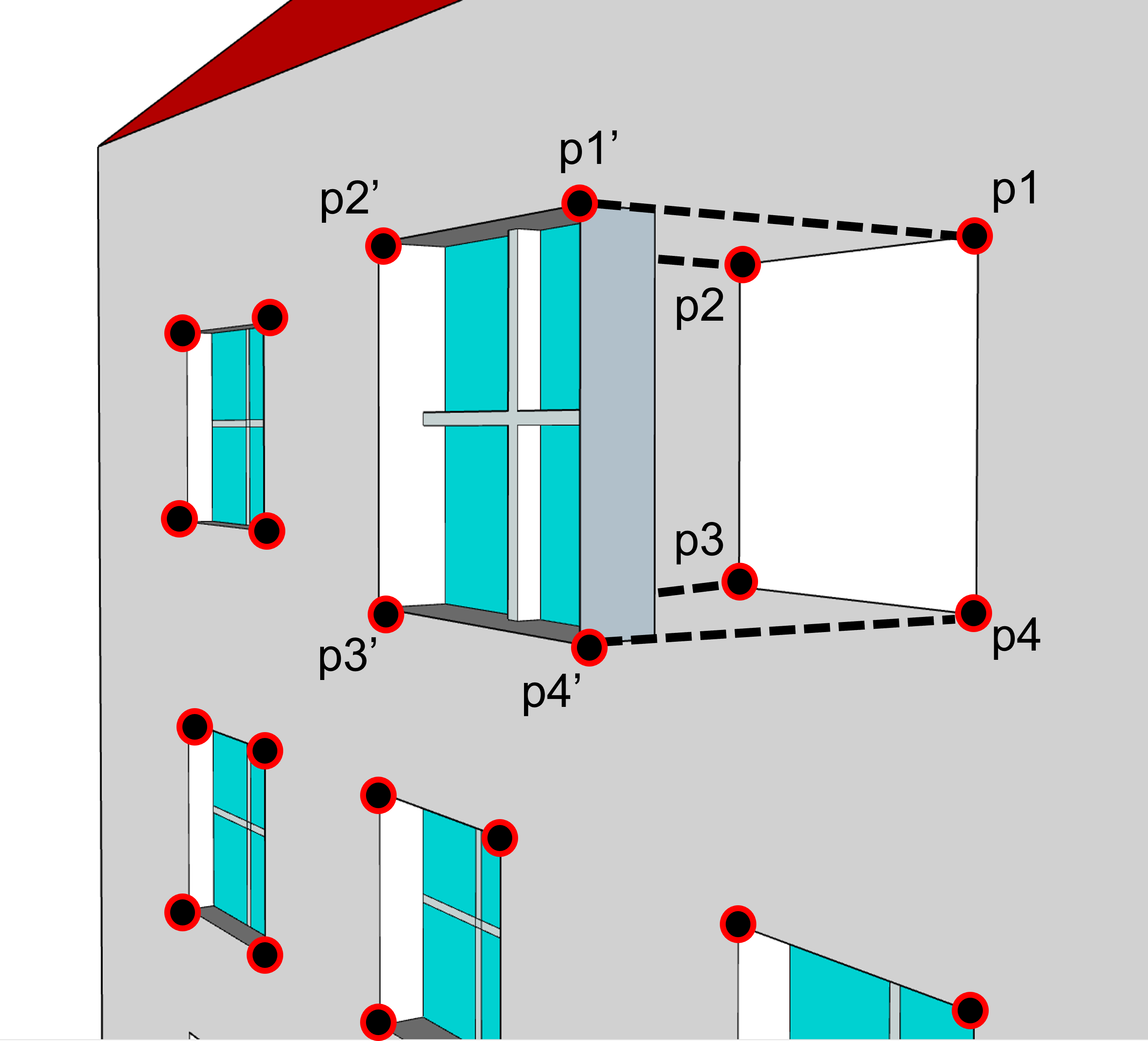}
\caption{Exemplary junction points (red-black) of cut shapes as fitting points on an example of pre-defined 3D window models} 
\label{holes}
\end{figure}

Regardless of the total building coverage, identified openings, such as underpasses, windows, and doors, shall cut out the Wall Surface geometry where required, and 3D geometries should be fitted into this empty space (Figure~\ref{holes}).
In this way, the highly-detailed representation, watertightness, and planarity of the prior model are maintained.

The notion of detection and reconstruction confidence is required since many applications nowadays rely not only on geometric accuracy but also on the associated confidence of the object's semantics.
For example, map-based navigation of cars uses a voting process fusing multiple sensor detections to decide upon the next manoeuvre~\cite{wilbers2019localization}.
Therefore, it is of great value to retain this information and add it as Generic Attribute called $confidence \in [0,1]$ to the specific object.
This attribute also allows for an update of the object's position and semantics when 
the next measurement epoch exceeds the confidence score~\cite{SesterZou}.
Here, of great value is the timestamp of acquisition, too~\cite{grogerOGCCityGeography2012}.

Although it is allowed to model an underpass at~\gls{LoD}2, in practice, it is rarely the case$^2$, primarily owing to the aerial perspective of data acquisition for~\gls{LoD}2 models~\cite{dukai2020generating}.
We deem underpasses as facade openings that substantially impact facade semantics and geometry; as such classify them as~\gls{LoD}3 features.

\begin{table}[htb]
    \caption{Analyzed point cloud and CityGML classes relevant for the LoD3 reconstruction with new, proposed functions (green) for absent ones~\cite{special_interest_group_3d_modeling_2020}}
    \footnotesize
    \centering
    \begin{tabular}{l crcccc} 
	\toprule
    Point cloud & CityGML & LoD & Function & Refinable & Confidence\\
    class & class &  &  &  & score \\
    \midrule
    ground surface  & Ground Surface & 1, 2, 3, 4 & - & $\widetilde{}$ & $\widetilde{}$ \\
    roof surface  & Roof Surface & 1, 2, 3, 4 & - & $\widetilde{}$ & $\widetilde{}$ \\
    wall & Wall Surface & 1, 2, 3, 4 & - & $\widetilde{}$ & \cmark \\
    window & Window & 3, 4 & - & \cmark & \cmark\\
    door & Door & 3, 4 & - & \cmark & \cmark\\
    underpass & Building Installation & 3, 4 & 1002 underpass & \cmark & \cmark\\
    balcony  & Building Installation & 3, 4 & 1000 balcony & \cmark & \cmark\\
    molding  & Building Installation & 3, 4 & \green{}{1016 molding} & \cmark & \cmark\\
    deco  & Building Installation & 3, 4 & \green{}{1017 deco} & \cmark & \cmark\\
    column  & Building Installation & 3, 4 & 1011 column & \cmark & \cmark\\
    arch  & Building Installation & 3, 4 & 1008 arch & \cmark & \cmark\\
    drainpipe  & Building Installation & 3, 4 & \green{}{1018 drainpipe} & \cmark & \cmark\\
    stairs  & Building Installation & 3, 4 & 1060 stairs & \cmark & \cmark\\
    blinds  & Building Installation & 3, 4 & \green{}{1019 blinds} & \cmark & \cmark\\
    \bottomrule
    \end{tabular}
    \label{codes}
\end{table}
\section{Examples}
\begin{figure}[h]
\includegraphics[width=\textwidth]{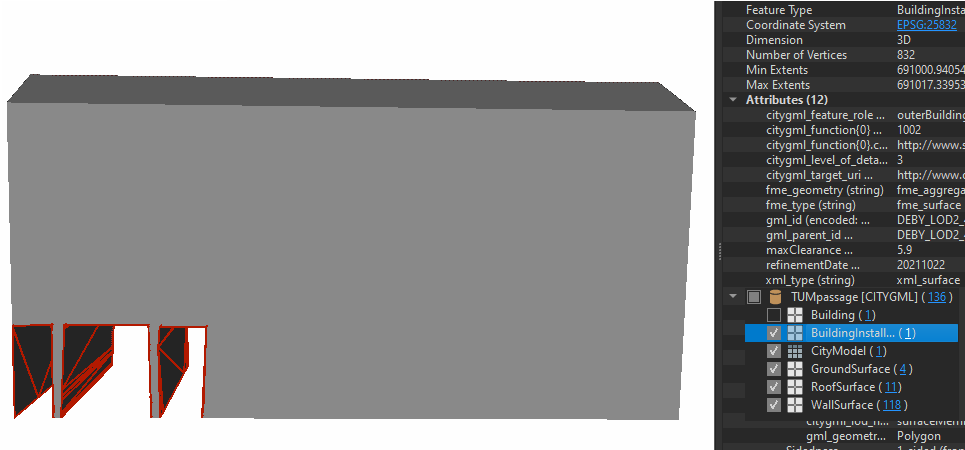}
\caption{Refinement result on an example of underpass embedded into the CityGML 2.0 model (right table)} 
\label{underpass}
\end{figure}
\begin{figure}[h]
\includegraphics[width=\textwidth]{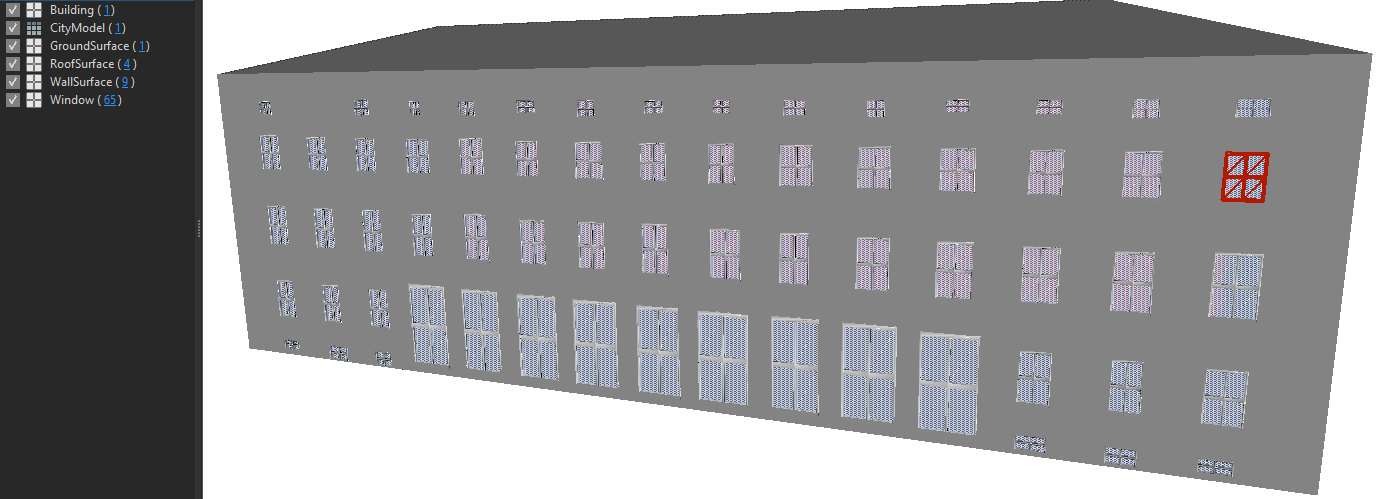}
\caption{Refinement result on an example of windows embedded into the CityGML 2.0 model (left table) with the highlighted one instance (red)} 
\label{window}
\end{figure}
\begin{figure}[h]
\includegraphics[width=\textwidth]{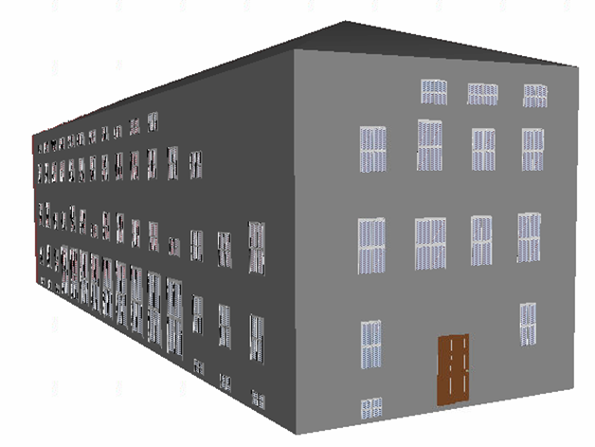}
\caption{Refinement result on an example of window and door types embedded into the CityGML 2.0 model. Note that here the TUM-MLS-2016 point cloud did not cover the whole building, and as such windows are partially missing, which can be complemented, for example, by an additional laser measurement or oblique images} 
\label{windowanddoor}
\end{figure}
The TUM-FAÇADE~\cite{tumfacadePaper} dataset, TUM-MLS-2016~\cite{zhu_tum-mls-2016_2020}, and openly available LoD2 and LoD3 building models~\footnote{https://github.com/tum-gis/tum2twin}~\cite{tumLoD2bavariaGov} depicting the main campus of the Technical University of Munich, Germany, were employed for the experiments.

The TUM-MLS-2016 was collected using the Mobile Distributed Situation Awareness (MODISSA) platform comprising two Velodyne HDL-64E LiDAR sensors.
The TUM-FAÇADE data was derived from the TUM-MLS-2016 and enriched by facade-level classes. 
The point cloud dataset was complemented by a proprietary point cloud dataset MF, which was also acquired at the Technical University of Munich campus. 
The point cloud was geo-referenced using a mobile mapping platform supported by the German SAPOS RTK system~\cite{mofa}.

The LoD2 building models were modeled by combining 2D footprints and aerial observations~\cite{RoschlaubBatscheider}. 
We textured the models by manually acquiring terrestrial pictures and projecting them to respective faces; we used a 13MP rear camera of a Xiaomi Redmi Note 5A smartphone and applied no orthorectification to mimic a direct projection from a mobile mapping platform or street vie imagery~\cite{biljeckiQualityStreetView}.
Based on LoD2 models and a combination of the TUM-FAÇADE and the proprietary MF point cloud, LoD3 models were manually created.
Such models served as a comparison set (ground-truth) to automatically generated models.
Additionally, they can be used as examples for practitioners. 
In this paper, we showcased the modeling paradigm of refinement; for the extensive discussion and comparison of the detection and reconstruction performance, see \cite{wysocki2023scan2lod3}, \cite{wysockiUnderpasses}, and \cite{wysockiVisibility}.

We present examples of three refinement types: windows, doors, and underpasses (Figure~\ref{underpass},~\ref{window}, and~\ref{windowanddoor});
The code is freely available within the repository~\footnote{[https://github.com/OloOcki/scan2lod3]}. 
\section{Discussion}
As shown in Figure~\ref{underpass},~\ref{window}, and~\ref{windowanddoor}, usage of planes of existing building models provides reliable projection targets for ray-to-model conflicts, semantic point clouds, and images; 
Moreover, it ensures homogeneous extent and pixel size.
Such plane-like assumptions have also proven effective for the co-registration of point clouds in the urban environment~\cite{tanjaSophiaRegistration,dong2020registration}.
In our experiments, we investigated terrestrial images projected to the plane, yet using pre-textured models or available oblique photos is possible, too, owing to the exploitation of the target plane.
More evidence shall improve performance as we base our inference on Bayesian reasoning. 
Therefore, combining other modalities is foreseeable and can mitigate occlusion issues as illustrated in Figure~\ref{windowanddoor}.  

Subsequent modeling also benefits from such a setup, as it is based on conventional constructive solid geometry operations, minimizing the reconstruction task complexity.
However, reliance on existing models requires flawless models and assumes their adherence to the CityGML standard.
On the other hand, in case the models do not exist for a particular region of interest, they can be generated using such established solutions as PolyFit~\cite{nan2017polyfit} for reconstructing LoD2 without footprints; or 3Dfier for prismatic generation of LoD1 building models~\cite{3dfier}.

In this experiment, we created a library of objects with merely one type per each facade element. 
However, works exist that tackle more sophisticated matching of various object library types while assuming correct detection, e.g., employing the Bag of Words approach~\cite{thomasFroechBoW}.
Note that our training dataset for point cloud semantic segmentation consists of several building types, which are representative of a central European architecture style and for such setup yields promising results~\cite{yuetanDeepLearning,wysocki2023scan2lod3}. 
The scalability for other architectural is foreseeable yet needs to be explored in future work.

\section{Conclusion}
According to the Cambridge Dictionary, the word refinement means "a small change that improves something" or "the process of making a substance pure"~\cite{cambridgeDictRefinement}.
In the case of our proposed strategy, both definitions are valid. On the one hand, we perform relatively minor changes to the 3D building model, improving its accuracy and semantic completeness significantly.
On the other hand, we preserve valid and remove undesired information from \gls{LoD}1 and 2 while reconstructing~\gls{LoD}3 building models.

We conclude that our refinement strategy allows for at-scale~\gls{LoD}3 building reconstruction by leveraging the ubiquity of existing semantic 3D building models and \gls{MLS} accuracy.
The experiments corroborate that the approach minimizes the reconstruction complexity and maintains 3D cadastre records by retaining 3D building database identifiers.  
This paper also presents guidelines for researchers and practitioners who investigate the reconstruction of~\gls{LoD}3 CityGML building models.
We firmly believe that the presented research can facilitate further development of algorithms for LoD3 reconstruction and unlock shown LoD3 building applications.
\section*{Acknowledgments}
This work was supported by the Bavarian State Ministry for Economic Affairs, Regional Development and Energy within the framework of the IuK Bayern project \textit{MoFa3D - Mobile Erfassung von Fassaden mittels 3D Punktwolken}, Grant No.\ IUK643/001.
Moreover, the work was conducted within the framework of the {Leonhard Obermeyer Center} at the Technical University of Munich (TUM).
We gratefully acknowledge the Geoinformatics team at the TUM for the valuable insights and for providing the CityGML datasets.
%
%
%
%
\bibliographystyle{splncs04}
\bibliography{mybiblio}
%




\end{document}